# Practical Linear Value-approximation Techniques for First-order MDPs

**Scott Sanner**
University of Toronto
Department of Computer Science
Toronto, ON, M5S 3H5, CANADA
ssanner@cs.toronto.edu

**Craig Boutilier**
University of Toronto
Department of Computer Science
Toronto, ON, M5S 3H5, CANADA
cebly@cs.toronto.edu

## Abstract

Recent work on approximate linear programming (ALP) techniques for first-order Markov Decision Processes (FOMDPs) represents the value function linearly w.r.t. a set of first-order basis functions and uses linear programming techniques to determine suitable weights. This approach offers the advantage that it does not require simplification of the first-order value function, and allows one to solve FOMDPs independent of a specific domain instantiation. In this paper, we address several questions to enhance the applicability of this work: (1) Can we extend the first-order ALP framework to approximate policy iteration and if so, how do these two algorithms compare? (2) Can we automatically generate basis functions and evaluate their impact on value function quality? (3) How can we decompose intractable problems with universally quantified rewards into tractable subproblems? We propose answers to these questions along with a number of novel optimizations and provide a comparative empirical evaluation on problems from the ICAPS 2004 Probabilistic Planning Competition.

## 1 Introduction

Markov decision processes (MDPs) have become the *de facto* standard model for decision-theoretic planning problems. While classic dynamic programming algorithms for MDPs require explicit state and action enumeration, recent techniques for exploiting propositional structure in factored MDPs [4] avoid explicit state and action enumeration. While such techniques for factored MDPs have proven effective, they cannot generally exploit first-order structure. Yet many realistic planning domains are best represented in first-order terms, exploiting the existence of domain objects, relations over these objects, and the ability to express objectives and action effects using quantification.

As a result, a new class of algorithms has been introduced to explicitly handle MDPs with relational (RMDP) and first-order (FOMDP) structure.[1] *Symbolic dynamic programming (SDP)* [5], *first-order value iteration (FOVIA)* [12, 13], and the *relational Bellman algorithm (ReBel)* [14] are model-based algorithms for solving FOMDPs and RMDPs, using appropriate generalizations of value iteration. *Approximate policy iteration* [7] induces rule-based policies from sampled experience in small-domain instantiations of RMDPs and generalizes these policies to larger domains. In a similar vein, *inductive policy selection using first-order regression* [9] uses regression to provide the hypothesis space over which a policy is induced. *Approximate linear programming (for RMDPs)* [10] is an approximation technique using linear program optimization to find a best-fit value function over a number of sampled RMDP domain instantiations.

A recent technique for first-order approximate linear programming (FOALP) [21] in FOMDPs approximates a value function by a linear combination of first-order basis functions. While FOALP incorporates elements of symbolic dynamic programming (SDP) [5], it uses a more compact approximation framework and avoids the need for logical simplification. This stands in contrast with exact value iteration frameworks [5, 12, 13, 14] that prove intractable in many cases due to blowup of the value function representation and the need to perform complex simplifications. And in contrast to approaches that require sampling of ground domains [7, 9, 10], FOALP solves a FOMDP at the *first-order level*, thus obviating the need for domain instantiation. However, FOALP is just one of many possible approaches to linear value approximation and this begs the question of whether we can generalize it to other approaches such as first-order approximate policy iteration (FOAPI). If so, it would be informative to obtain a comparative empirical evaluation of these algorithms.

However, determining the most effective algorithm for linear value-approximation is only the first step towards the development of practical approximation techniques for FOMDPs. We address the important issue of automatic basis function generation by extending regression-based techniques originally proposed by Gretton and Thiebaux [9]. In addition, we address issues that arise with universal rewards—while symbolic dynamic programming is well-defined for FOMDP domains with universal rewards, classical first-order logic is insufficient for defining a compact set of basis functions that can adequately approximate the value function in such domains. We propose a technique for *decomposing* problems with universal rewards while facilitating "coordination" among decomposed subproblems. Finally, we present a number of novel optimizations that enhance the performance of FOALP and FOAPI. We provide a comparative evaluation of these algorithms on problems from the ICAPS 2004 International Probabilistic Planning Competition (IPPC).

---

[1]We use the term *relational MDP* to refer models that allow implicit existential quantification, and *first-order MDP* for those with explicit existential and universal quantification.

## 2 Markov Decision Processes

We first review linear value-approximation of MDPs.

### 2.1 MDP Representation

An MDP consists of: a finite state space $S$; a finite set of actions $A$; a transition function $T$, with $T(s, a, \cdot)$ denoting a distribution over $S$ for all $s \in S, a \in A$; and a reward function $R : S \times A \to \mathbb{R}$. Our goal is to find a policy $\pi$ that maximizes the value function, defined using the infinite horizon, discounted reward criterion: $V^\pi(s) = E_\pi[\sum_{t=0}^\infty \gamma^t \cdot r^t | s]$, where $r^t$ is a reward obtained at time $t$ and $0 \leq \gamma < 1$ is the discount factor.

For any function $V$ over $S$ and policy $\pi$, define the *policy backup* operator $B^\pi$ as: $(B^\pi V)(s) = \gamma \sum_t T(s, \pi(s), t) V(t)$. The *action backup* operator $B^a$ for action $a$ is: $(B^a V)(s) = \gamma \sum_t T(s, a, t) V(t)$. The function $V^\pi(s)$ satisfies the fixed point relationship $V^\pi(s) = R(s, \pi(s)) + (B^\pi V^\pi)(s)$. Furthermore, the Q-function $Q^\pi$, denoting the expected future discounted reward achieved by taking action $a$ in state $s$ and following policy $\pi$ thereafter, satisfies $Q^\pi(s, a) = R(s, a) + (B^a V^\pi)(s)$. We define the *greedy policy* w.r.t. $V$ as follows: $\pi_{gre}(V)(s) = \arg\max_a R(s, a) + (B^a V)(s)$. If $\pi^*$ is the optimal policy and $V^*$ its value function, we have the relationship $V^*(s) = \max_a R(s, a) + (B^a V^*)(s)$. Letting $Q^*(s, a) = R(s, a) + (B^a V^*)(s)$, we also have $\pi^*(s) = \pi_{gre}(V^*)(s) = \arg\max_a Q^*(s, a)$.

### 2.2 Linear Value Approximation for MDPs

In a linear value function representation, we represent $V$ as a linear combination of $k$ basis functions $b_j$: $V(s) = \sum_{j=1}^k w_j b_j(s)$. Our goal is to find weights that approximate the optimal value function as closely as possible. We note that because both backup operators $B^\pi$ and $B^a$ are linear operators, the backup of a linear combination of basis functions is just the linear combination of the backups of the individual basis functions.

**Approximate Linear Programming:** One way of finding a good linear approximation is to cast the optimization problem as a linear program that directly solves for the weights of an $L_1$-minimizing approximation of the optimal value function [6]:

Variables: $w_1, \ldots, w_k$

$$\text{Minimize:} \sum_{s \in S} \sum_{j=1}^k w_j b_j(s) \quad (1)$$

Subject to: $0 \geq R(s,a) + \sum_{j=1}^k w_j[(B^a b_j)(s) - b_j(s)]$ ; $\forall a, s$

While the size of the objective and the number of constraints in this LP is proportional to the number of states (and therefore exponential), recent solution techniques use compact, factored basis functions and exploit the reward and transition structure of factored MDPs [11, 22], making it possible to avoid generating an exponential number of constraints (and rendering the entire LP compact).

**Approximate Policy Iteration:** Likewise, we can generalize policy iteration to the approximate case by calculating successive iterations of weights $w_j^{(i)}$ that represent the best approximation of the fixed point value function for policy $\pi^{(i)}$ at iteration $i$. We do this by performing the following two steps at each iteration: (1) deriving the greedy policy: $\pi^{(i+1)} \leftarrow \pi_{gre}(\sum_{j=1}^k w_j^{(i)} b_j(s))$ and (2) using the following LP to determine the weights for the Bellman-error-minimizing approximate value function:

Variables: $w_1^{(i+1)}, \ldots, w_k^{(i+1)}$

Minimize: $\phi^{(i+1)}$ (2)

Subject to: $\phi^{(i+1)} \geq \left| R(s, \pi(s)) + \sum_{j=1}^k [w_j^{(i+1)}(B^{\pi^{(i+1)}} b_j)(s)] - \sum_{j=1}^k [w_j^{(i+1)} b_j(s)] \right|$ ; $\forall a, s$

If policy iteration converges (i.e., if $\vec{w}^{(i+1)} = \vec{w}^{(i+1)}$), then Guestrin *et al.* [11] provide the following bound on the loss of $V^{(i+1)}$ w.r.t. the optimal value function $V^*$ (since the Bellman error is bounded by the objective $\phi^{(i+1)}$ of the optimal LP solution at iteration $i+1$):

$$\|V^* - V^{(i+1)}(s)\|_\infty \leq \frac{2\gamma \phi^{(i+1)}}{1 - \gamma} \quad (3)$$

## 3 First-Order MDPs

### 3.1 The Situation Calculus

The situation calculus [19] is a first-order language for axiomatizing dynamic worlds. Its basic ingredients consist of *actions*, *situations*, and *fluents*. Actions are first-order terms involving action function symbols. For example, the action of driving a truck $t$ from city $c_1$ to city $c_2$ might be denoted by the action term $drive(t, c_1, c_2)$. A situation is a first-order term denoting the occurrence of a sequence of actions. These are represented using a binary function symbol $do$: $do(\alpha, s)$ denotes the situation resulting from doing action $\alpha$ in situation $s$. In a logistics domain, the situation term $do(drive(t, c_2, c_3), do(drive(t, c_1, c_2), S_0))$ denotes the situation resulting from executing sequence $[drive(t, c_1, c_2), drive(t, c_2, c_3)]$ in $S_0$. Relations whose truth values vary between states are called fluents, and are denoted by predicate symbols whose last argument is a situation term. For example, $TAt(t, paris, s)$ is a relational fluent meaning that truck $t$ is in $paris$ in situation $s$.[2]

A domain theory is axiomatized in the situation calculus with four classes of axioms [19]. The most important of these are *successor state axioms (SSAs)*. There is one SSA for each fluent $F(\vec{x}, s)$, with syntactic form $F(\vec{x}, do(a, s)) \equiv \Phi_F(\vec{x}, a, s)$ where $\Phi_F(\vec{x}, a, s)$ is a formula with free variables among $a, s, \vec{x}$. These characterize the truth values of the fluent $F$ in the next situation $do(a, s)$ in terms of the current situation $s$, and embody a solution to the frame problem for deterministic actions [19].

The *regression* of a formula $\psi$ through an action $a$ is a formula $\psi'$ that holds prior to $a$ being performed iff $\psi$ holds after $a$. We refer the reader to [5, 21] for a formal definition and discussion of the $Regr(\cdot)$ operator. Here we simply note that it is defined compositionally and that regression of a formula reduces to the regression of all fluents in a way that is naturally supported by the format of the SSAs.

### 3.2 Case Representation and Operators

Prior to generalizing the situation calculus to permit a first-order representation of MDPs, we introduce a *case notation* to allow first-order specifications of the rewards, probabilities, and values required for FOMDPs (see [5, 21] for formal details):

$$t = \begin{array}{|c|} \hline \phi_1 : t_1 \\ \hline \vdots \; : \; \vdots \\ \hline \phi_n : t_n \\ \hline \end{array} \equiv \bigvee_{i \leq n}\{\phi_i \wedge t = t_i\}$$

Here the $\phi_i$ are *state formulae* (whose situation term does not use $do$) and the $t_i$ are terms. Often the $t_i$ will be constants and the $\phi_i$ will partition state space. For example, using $Dst(t, c)$ to indicate the destination of truck $t$ is city $c$, we may represent our reward function $rCase(s, a)$ as:

$$rCase(s,a) = \begin{array}{|ll|} \hline a = noop \wedge \forall t, c \; TAt(t,c,s) \supset Dst(t,c) & : 10 \\ \hline a \neq noop \wedge \forall t, c \; TAt(t,c,s) \supset Dst(t,c) & : \;\; 9 \\ \hline \exists t, c \; TAt(t,c,s) \wedge \neg Dst(t,c) & : \;\; 0 \\ \hline \end{array}$$

Here, we receive a reward of 10 (9) if all trucks are at their destination and a noop is (not) performed. In all other cases we receive 0 reward. We use $vCase(s)$ to represent value functions in exactly the same manner.

---
[2]In contrast to states, situations reflect the entire history of action occurrences. However, the specification of dynamics is Markovian and allows recovery of state properties from situation terms.

Intuitively, to perform an operation on two case statements, we simply take the cross-product of their partitions and perform the corresponding operation on the resulting paired partitions. Letting each $\phi_i$ and $\psi_j$ denote generic first-order formulae, we can perform the "cross-sum" $\oplus$ of two case statements in the following manner:

$$\begin{array}{|c|} \hline \phi_1 : 10 \\ \hline \phi_2 : 20 \\ \hline \end{array} \oplus \begin{array}{|c|} \hline \psi_1 : 1 \\ \hline \psi_2 : 2 \\ \hline \end{array} = \begin{array}{|c|} \hline \phi_1 \wedge \psi_1 : 11 \\ \hline \phi_1 \wedge \psi_2 : 12 \\ \hline \phi_2 \wedge \psi_1 : 21 \\ \hline \phi_2 \wedge \psi_2 : 22 \\ \hline \end{array}$$

Likewise, we can perform $\ominus$ and $\otimes$ by, respectively, subtracting or multiplying partition values (as opposed to adding them) to obtain the result. Some partitions resulting from the application of the $\oplus$, $\ominus$, and $\otimes$ operators may be inconsistent; we simply discard such partitions since they can obviously never correspond to any world state.

We define four additional operators on cases [5, 21]: $Regr(\cdot)$, $\exists \vec{x}$, max, and $\cup$. Regression $Regr(C)$ and existential quantification $\exists \vec{x} C$ can both be applied directly to the individual partition formulae $\phi_i$ of case $C$. The maximization operation $\max C$ sorts the partitions of case $C$ from largest to smallest, rendering them disjoint in a manner that ensures each portion of state space is assigned the highest value. The union operation $C_1 \cup C_2$ denotes a simple union of the case partitions from cases $C_1$ and $C_2$.

### 3.3 Stochastic Actions and the Situation Calculus

To generalize the classical situation calculus to stochastic actions required by FOMDPs, we decompose stochastic "agent" actions into a *collection* of deterministic actions, each corresponding to a possible outcome of the stochastic action. We then specify a distribution according to which "nature" may choose a deterministic action from this set whenever that stochastic action is executed. As a consequence we need only formulate SSAs using the deterministic *nature's choices* [1, 5], thus obviating the need for a special treatment of stochastic actions in SSAs.

Letting $A(\vec{x})$ be a stochastic action with nature's choices (i.e., deterministic actions) $n_1(\vec{x}), \cdots, n_k(\vec{x})$, we represent the distribution over $n_i(\vec{x})$ given $A(\vec{x})$ using the notation $pCase(n_j(\vec{x}), A(\vec{x}), s)$. Continuing our logistics example, if the effect of driving a truck depends on whether it is snowing in the city of origin, then we decompose the stochastic $drive$ action into two deterministic actions $driveS$ and $driveF$, denoting success and failure, respectively, and specify a distribution over nature's choice:

$$pCase(\; driveS(t, c_1, c_2), \\ drive(t, c_1, c_2), s) \;=\; \begin{array}{|c|} \hline snow(c_1, s) : 0.6 \\ \hline \neg snow(c_1, s) : 0.9 \\ \hline \end{array}$$

$$pCase(\; driveF(t, c_1, c_2), \\ drive(t, c_1, c_2), s) \;=\; \begin{array}{|c|} \hline snow(c_1, s) : 0.4 \\ \hline \neg snow(c_1, s) : 0.1 \\ \hline \end{array}$$

Next, we define the SSAs in terms of these deterministic choices.[3] Assuming that nature's choice of deterministic actions for stochastic action $drive(t, c_1, c_2)$ decomposes as above, we can express an SSA for $TAt$:

$TAt(t, c, do(a, s)) \equiv$
$\quad \exists c_1\ TAt(t, c_1, s) \land a = driveS(t, c_1, c) \lor$
$\quad TAt(t, c, s) \land \neg(\exists c_2\ c \neq c_2 \land a = driveS(t, c, c_2))$

Intuitively, the only actions that can change the fluent $TAt$ are successful *drive* actions. If a successful *drive* action does not occur then the fluent remains unchanged.

### 3.4 Symbolic Dynamic Programming

Backing up a value function $vCase(s)$ through an action $A(\vec{x})$ yields a case statement containing the logical description of states that would give rise to $vCase(s)$ after doing action $A(\vec{x})$, as well as the values thus obtained (i.e., a $Q(s, a)$ function in classical MDPs). There are in fact *three* types of backups that we can perform. The first, $B^{A(\vec{x})}$, regresses a value function through an action and produces a case statement with *free variables* for the action parameters. The second, $B^A$, existentially quantifies over the free variables $\vec{x}$ in $B^{A(\vec{x})}$. The third, $B^A_{\max}$ applies the max operator to $B^A$ which results in a case description of the regressed value function indicating the best value that could be achieved by executing *any* instantiation of $A(\vec{x})$ in the pre-action state. To define the backup operators, we first define a slightly modified version of the *first-order decision theoretic regression (FODTR)* operator [5]:

$FODTR(vCase(s), A(\vec{x})) =$
$\quad \gamma\ [\oplus_j \{pCase(n_j(\vec{x}), s) \otimes Regr(vCase(do(n_j(\vec{x}), s)))\}]$

We next next define the three backup operators:

$$B^{A(\vec{x})}(vCase(s)) = rCase(s, a) \oplus FODTR(vCase(s), A(\vec{x})) \quad (4)$$

$$B^A(vCase(s)) = \exists \vec{x}\ B^{A(\vec{x})}(vCase(s)) \quad (5)$$

$$B^A_{\max}(vCase(s)) = \max(B^A(vCase(s))) \quad (6)$$

Previous work [21] provides examples of $B^{A(\vec{x})}$ and $B^A_{\max}$.

## 4 Linear Value Approximation for FOMDPs

### 4.1 Value Function Representation

Following [21], we represent a value function as a weighted sum of $k$ *first-order basis functions*, denoted $bCase_i(s)$, each containing a *small* number of formulae that provide a first-order abstraction of state space:

$$vCase(s) = \oplus_{i=1}^{k}\ w_i \cdot bCase_i(s) \quad (7)$$

Using this format, we can often achieve a reasonable approximation of the exact value function by exploiting the

---

[3]SSAs can often be compiled from "effect" axioms that specify action effects [19] and effect axioms can be compiled from PPDDL probabilistic planning domain specifications [25].

---

additive structure inherent in many real-world problems (e.g., additive reward functions or problems with independent subgoals). Unlike exact solution methods where value functions can grow exponentially in size during the solution process and must be logically simplified [5], here we maintain the value function in a compact form that requires no simplification, just discovery of good weights.

We can easily apply the backup operator $B^A$ to this representation and obtain some simplification as a result of the structure in Eq. 7. We simply substitute the value function expression in Eq. 7 into the definition $B^{A(\vec{x})}$ (Eq. 4). Exploiting the properties of the $Regr$ and $\oplus$ operators, we find that the backup $B^{A(\vec{x})}$ of a linear combination of basis functions is simply the linear combination of the FODTR of each basis function:

$$B^{A(\vec{x})}(\oplus_i w_i bCase_i(s)) = \quad (8)$$
$$rCase(s, a) \oplus (\oplus_i w_i FODTR(bCase_i(s), A(\vec{x})))$$

Given the definition of $B^{A(\vec{x})}$ for a linear combination of basis functions, corresponding definitions of $B^A$ and $B^A_{\max}$ follow directly from Eqs. 5 and 6. It is important to note that during the application of these operators, we never explicitly ground states or actions, in effect achieving *both state and action space abstraction*.

### 4.2 First-order Approximate Linear Programming

Now we have all of the building blocks required to define first-order approximate linear programming (FOALP) and first-order approximate policy iteration (FOAPI). For now we simply focus on the algorithm definitions; we will address efficient implementation in a subsequent section.

FOALP was introduced by Sanner and Boutilier [21]. Here we present a linear program (LP) with first-order constraints that generalizes Eq. 1 from MDPs to FOMDPs:

Variables: $w_i\ ;\ \forall i \leq k$

Minimize: $\sum_{i=1}^{k} w_i \sum_{\langle \phi_j, t_j \rangle \in bCase_i} \frac{t_j}{|bCase_i|}$

Subject to: $0 \geq B^A_{\max}(\oplus_{i=1}^{k} w_i \cdot bCase_i(s))$
$\quad \ominus (\oplus_{i=1}^{k} w_i \cdot bCase_i(s))\ ;\ \forall A, s \quad (9)$

The objective of this LP requires some explanation. If we were to directly generalize the objective for MDPs to that of FOMDPs, the objective would be ill-defined (it would sum over infinitely many situations). To remedy this, we suppose that each basis function partition is chosen because it represents a potentially useful partitioning of state space, and thus sum over each case *partition*.

This LP also contains a first-order specification of constraints, which somewhat complicates the solution. Before tackling this, we introduce a general *first-order LP* format

that we can reuse for FOAPI:

$$\begin{aligned}
&\text{Variables: } v_1, \ldots, v_k \text{ ;} \\
&\text{Minimize: } f(v_1, \ldots, v_k) \\
&\text{Subject to: } 0 \geq case_{1,1}(s) \oplus \ldots \oplus case_{1,n}(s) \text{ ; } \forall s \\
&\qquad\qquad\qquad \vdots \\
&\qquad\qquad 0 \geq case_{m,1}(s) \oplus \ldots \oplus case_{m,n}(s) \text{ ; } \forall s
\end{aligned} \quad (10)$$

The variables and objective are as defined in a typical LP, the main difference being the form of the constraints. While there are an infinite number of constraints (i.e., one for every situation $s$), we can work around this since case statements are finite. Since the value $t_i$ for each case partition $\langle \phi_i(s), t_i \rangle$ is piecewise constant over all situations satisfying $\phi_i(s)$, we can explicitly sum over the $case_i(s)$ statements in each constraint to yield a single case statement. For this "flattened" case statement, we can easily verify that the constraint holds in the finite number of piecewise constant partitions of the state space. However, generating the constraints for each "cross-sum" can yield an exponential number of constraints. Fortunately, we can generalize constraint generation techniques [22] to avoid generating all constraints. We refer to [21] for further details. Taken together, these techniques yield a practical FOALP solution to FOMDPs.

### 4.3 First-order Approximate Policy Iteration

We now turn to the first contribution of this paper, a novel generalization of approximate policy iteration from the classical MDP case (Eq. 1) to FOMDPs.

Policy iteration requires a suitable first-order policy representation. Given a value function $vCase(s)$ it is easy to derive a greedy policy from it. Assuming we have $m$ parameterized actions $\{A_1(\vec{x}), \ldots, A_m(\vec{x})\}$, we can represent the policy $\pi Case(s)$ as:

$$\pi Case(s) = \max_{i=1\ldots m} (\bigcup B^{A_i}(vCase(s))) \quad (11)$$

Here, $B^{A_i}(vCase(s))$ represents the values that can be achieved by any instantiation of the action $A_i(\vec{x})$. The max case operator enforces that each portion of pre-action state space is assigned the maximal Q-function partition. For bookkeeping purposes, we require that each partition $\langle \phi, t \rangle$ in $B^{A_i}(vCase(s))$ maintain a mapping to the action $A_i$ that generated it, which we denote as $\langle \phi, t \rangle \to A_i$. Then, given a particular world state $s$, we can evaluate $\pi Case(s)$ to determine which policy partition $\langle \phi, t \rangle \to A_i$ is satisfied in $s$ and thus, which action $A_i$ should be applied. If we retrieve the bindings of the existentially quantified action variables in $\phi$ (recall that $B^{A_i}$ existentially quantifies these), we can easily determine the parameterization of action $A_i$ that should apply according to the policy.

For our algorithms, it is useful to define a set of case statements for each action $A_i$ that is satisfied only in the world states where $A_i$ should be applied according to $\pi Case(s)$. Consequently, we define an action restricted policy $\pi Case_{A_i}(s)$ as follows:

$$\pi Case_{A_i}(s) = \{\langle \phi, t \rangle | \langle \phi, t \rangle \in \pi Case(s) \text{ and } \langle \phi, t \rangle \to A_i\}$$

Following the approach to approximate policy iteration for factored MDPs provided by Guestrin *et al.* [11], we can generalize approximate policy iteration to the first-order case by calculating successive iterations of weights $w_j^{(i)}$ that represent the best approximation of the fixed point value function for policy $\pi Case^{(i)}(s)$ at iteration $i$. We do this by performing the following two steps at every iteration $i$: (1) Obtaining the policy $\pi Case(s)$ from the current value function and weights ($\sum_{j=1}^k w_j^{(i)} bCase_j(s)$) using Eq. 11, and (2) solving the following LP in the format of Eq. 10 that determines the weights of the Bellman-error-minimizing approximate value function for policy $\pi Case(s)$:

Variables: $w_1^{(i+1)}, \ldots, w_k^{(i+1)}$

Minimize: $\phi^{(i+1)}$  (12)

Subject to: $\phi^{(i+1)} \geq \Big| \pi Case_A(s) \oplus \oplus_{j=1}^k [w_j^{(i+1)} bCase_j(s)]$

$\ominus \oplus_{j=1}^k w_j^{(i+1)} (B_{\max}^A bCase_j)(s) \Big|; \forall A, s$

We've reached convergence *if* $\pi^{(i+1)} = \pi^{(i)}$. If the policy iteration converges, then we note that the loss bounds for API (Eq. 3) generalize directly to the first-order case.

## 5 Greedy Basis Function Generation

The use of linear approximations requires a good set of basis functions that span a space that includes a good approximation to the value function. While some work has addressed the issue of basis function generation [18, 16], none has been applied to RMDPs or FOMDPs. We consider a basis function generation method that draws on the work of Gretton and Thiebaux [9], who use inductive logic programming (ILP) techniques to construct a value function from sampled experience. Specifically, they use regressions of the reward as candidate building blocks for ILP-based construction of the value function. This technique has allowed them to generate fully or $k$-stage-to-go optimal policies for a range of Blocks World problems.

We leverage a similar approach for generating candidate basis functions for use in the FOALP or FOAPI solution techniques. Fig. 1 provides an overview of our basis function generation algorithm. The motivation for this approach is as follows: if some portion of state space $\phi$ has value $v > \tau$ in an existing approximate value function for some nontrivial threshold $\tau$, then this suggests that states that can reach this region (i.e., found by $Regr(\phi)$ through some action) should also have reasonable value. However, since we have already assigned value to $\phi$, we want the new basis function to focus on the area of state space not covered by $\phi$ so we negate it and conjoin it with $Regr(\phi)$. This "orthogonality" of newly generated basis functions also allows for computation optimizations (see Sec. 7).

> **Input:** A first-order MDP specification, a value threshold $\tau$, an iteration limit $n$, and a solution method (FOALP or FOAPI).
> **Output:** Weights $w_i$ and basis functions $bCase_i(s)$ for an approximated value function containing all regressions of $rCase(s)$ having value at least value $\tau$.
>
> 1. Begin with the reward $rCase(s)$ as the initial candidate basis function. (We note that $rCase(s)$ can be a sum of cases, so we can start with many basis functions.)
> 2. For every basis function partition $\langle \phi_i(s), t_i \rangle$ and action $A_i$, derive $\neg \phi_i \wedge \exists \vec{x}\ Regr(\phi_i(do(A_i(\vec{x}), s)))$, adding a new basis function consisting of it and its negation having respective values 1 and 0. (Redundant basis functions are not inserted.)
> 3. Solve for the weights $w_i$ using FOALP or FOAPI.
> 4. If the weight of any basis function is below a threshold $\tau$, discard the partition and ensure that it is not regenerated in the future.
> 5. If new basis functions were generated on this step and the iteration limit $n$ has not been reached, return to step 2.

Figure 1: The basis function generation algorithm.

# 6 Handling Universal Rewards

In first-order domains, we are often faced with *universal reward expressions* that assign some positive value to the world states satisfying a formula of the general form $\forall y\ \phi(y, s)$, and 0 otherwise. For instance, in our logistics example a reward may be given for having all trucks at their assigned destination: $\forall t, cDst(t, c) \rightarrow TAt(t, c, s)$. One difficulty with such rewards is that our basis function approach provides a piecewise constant approximation to the value function (i.e., each basis function aggregates state space into regions of equal value, with the linear combination simply providing constant values over somewhat smaller regions). However, the value function for problems with universal rewards typically depends (often in a linear or exponential way) on the *number* of domain objects of interest. For instance, in our example, value at a state depends on the number of trucks not at their proper destination (since that impacts the time it will take to obtain the reward). Unfortunately, this cannot be represented concisely using the piecewise constant decomposition offered by first-order basis functions. As noted by Gretton and Thiebaux [9], effectively handling universally quantified rewards is one of the most pressing issues in the practical solution of FOMDPs.

To address this problem we adopt a decompositional approach, motivated in part by techniques for additive rewards in MDPs [3, 23, 17, 18]. Intuitively, given a goal-oriented reward that assigns positive reward if $\forall y G(y, s)$ is satisfied, and zero otherwise, we can decompose it into a set of ground goals $\{G(\vec{y_1}), \ldots, G(\vec{y_n})\}$ for all possible $\vec{y_j}$ in a ground domain of interest. If we reach a state where all ground goals are true, then we have satisfied $\forall y G(y, s)$.

Of course, our methods solve FOMDPs without knowledge

> **Input:** (1) For each action template $A_i(\vec{x})$, a set of Q-functions $Q_{G(\vec{y^*})}(A_i(\vec{x}), s)$ for a specific ground instantiation $\vec{y^*}$ of a goal $G$. (2) A set of $n$ unsatisfied goals $\{G(\vec{y_1}), \ldots, G(\vec{y_n})\}$ to achieve. (3) A ground state $s$ to find the best action for.
> **Output:** The optimal ground action $A(\vec{x^*})$ to execute w.r.t. to the given state and additive decomposition of unsatisfied goals: $A(\vec{x^*}) = \arg\max_{i,\vec{x}} \sum_{j=1}^{n} Q_{G(\vec{y_j})}(A_i(\vec{x}), s)$
>
> 1. For each action template $A_i(\vec{x})$ and goal $G(\vec{y_j})$, *replace* all occurrences of $\vec{y^*}$ in $Q_{G(\vec{y^*})}(A_i(\vec{x}), s)$ with $\vec{y_j}$ to obtain a set of *goal-specific* Q-functions for each $A_i(\vec{x})$: $\{Q_{G(\vec{y_1})}(A_i(\vec{x}), s), \ldots, Q_{G(\vec{y_n})}(A_i(\vec{x}), s)\}$.
> 2. Initialize an empty hash table $h$ whose entries $A(\vec{x}) \rightarrow v$ map ground actions $A(\vec{x})$ to their corresponding value $v$.
> 3. For $j = 1 \ldots n$ do:
>    For all $A_i$ and case partitions $p$ in $Q_{G(\vec{y_j})}(A_i(\vec{x}), s)$:
>    - *NOTE: By construction from the $B^A$ operator, a case partition $p$ for a Q-function $Q_{G(\vec{y_j})}(A_i(\vec{x}), s)$ has the format $\langle \exists \vec{x}\ \phi(x) : t \rangle$ where we obtain the value $t$ when $A_i(\vec{x})$ is performed if the ground binding $\vec{x}$ is one satisfying $\exists \vec{x}\ \phi(x)$.*
>    - For each ground binding $\vec{x}$ satisfying $\exists x\ \phi(x)$:
>      – If $A(x) \rightarrow v$ is already in $h$ then: update $h$ to contain $A(x) \rightarrow (v + \frac{t}{n})$.
>      – Else: update $h$ to contain $A(x) \rightarrow \frac{t}{n}$.

Figure 2: Policy evaluation algorithm for universal rewards.

of the specific domain, so the set of ground goals that will be faced at run-time is unknown. So in the offline solution of the MDP we assume a *generic* ground goal $G(\vec{y^*})$ for a "generic" object vector $\vec{y^*}$. It is easy to construct an instance of the reward function $rCase(s)$ for this single goal, and solve for this simplified generic goal using FOALP or FOAPI. This produces a value function and policy that assumes that $\vec{y^*}$ is the only object vector of interest (i.e., satisfying relevant type constraints and preconditions) in the domain. From this, we can also derive the optimal Q-function for the simplified "generic" domain (and action template $A_i(\vec{x})$): $Q_{G(\vec{y^*})}(A_i(\vec{x}), s) = B^{A_i}(vCase(s))$.[4] Intuitively, given a ground state $s$, the optimal action for this generic goal can be determined by finding the ground $A_i(\vec{x^*})$ for this $s$ with max Q-value.

With the solution (i.e., optimal Q-function) of a generic goal FOMDP in hand, we address the online problem of action selection for a specific domain instantiation. Assume a set of ground goals $\{G(\vec{y_1}), \ldots, G(\vec{y_n})\}$ corresponding to a specific domain given at run-time. If we assume that (typed) domain objects are treated uniformly in the uninstantiated FOMDP, as is the case in many logistics and planning problems, then we obtain the Q-function for any goal $G(\vec{y_j})$ by replacing all ground terms $\vec{y^*}$ with the respective terms $\vec{y_j}$ in $Q_{G(\vec{y^*})}(A_i(\vec{x}), s)$ to ob-tain $Q_{G(\vec{y_j})}(A_i(\vec{x}), s)$.

---

[4]Since the $B^A$ operator can often retain much of the additive structure in the linear approximation of $vCase(s)$ [21], representation and computation with this Q-function is *very* efficient.

Action selection requires finding an action that maximizes value w.r.t. the original universal reward. Following [3, 17], we do this by treating the *sum of the Q-values* of any action in the subgoal MDPs as a measure of its Q-value in the joint (original) MDP. Specifically, we assume that each goal contributes uniformly and additively to the reward, so the Q-function for an entire set of ground goals $\{G(\vec{y_1}), \ldots, G(\vec{y_n})\}$ determined by our domain instantiation is just $\sum_{j=1}^{n} \frac{1}{n} Q_{G(\vec{y_j})}(A_i(\vec{x}), s)$. Action selection (at run-time) in any ground state is realized by choosing that action with maximum joint Q-value. Naturally, we do not want to explicitly create the joint Q-function, but instead use an efficient scoring technique that evaluates potentially useful actions by iterating through the individual Q-functions as described in Fig. 2. While this additive and uniform decomposition may not be appropriate for all domains with goal-oriented universal rewards, we have found it to be highly effective for the two domains examined in this paper. And while this approach can only currently handle rewards with universal quantifiers, this reflects the form of many planning problems. Nonetheless, there are potential extensions of this technique for more complex universal rewards, the general question being how to assign credit among the constituents of such a reward.

## 7 Optimizations

Following are a few novel additional optimizations that provided substantial performance enhancements of our FOALP and FOAPI implementations. First, and most importantly, the style of generating orthogonal basis functions in Fig. 1 has some very nice computational properties that we can exploit. In short, when searching for the maximum partition among $n$ disjoint basis functions, one need only consider taking 1 of $n$ *true* partitions (each basis function has $\phi$ and $\neg\phi$) setting the other $n-1$ basis functions to its *false* partition. Clearly any other setting would result in an inconsistent state due to the disjointness of the $n$ basis functions. Consequently, the search for a consistent state reduces from an exponential complexity of $2^n$ combinations down to a polynomial complexity of $n$ combinations (trying each *true* partition of a basis function in turn).

Second, one can replace the $B^A_{\max}$ operators in the *constraints* for FOALP and FOAPI with the much simpler $B^A$ operator that does not introduce the blowup that occurs from enforcing disjointness in the $B^A_{\max}$ operator. Since we know that we only use the constraints when searching for a max (i.e., during constraint generation [21]), the max over $B^A$ will implicitly enforce the max constraint of $B^A_{\max}$. While we omit a proof, it is straightforward to show that the maximal value and therefore the maximally violated constraint that we need during constraint generation is the same whether we use $B^A_{\max}$ or $B^A$.

Third, while first-order simplification techniques are not required to perform FOALP or FOAPI, some simplification can save a substantial amount of theorem proving time. We used a simple BDD-based simplification technique as follows: Given a first-order formula, we rearrange and push quantifiers down into subformulae as far as possible. This exposes a propositional super-structure (very common in FOMDP problems) that can be mapped into a BDD. This BDD structure is useful because it reduces *propositional* redundancy in the formula representation by removing duplicate or inconsistent closed first-order formulae that are repeated frequently due to the naive conjunction of the case operators (mainly $\oplus$, $\ominus$, and $\otimes$).

## 8 Empirical Results

We applied FOALP and FOAPI to the *Box World* logistics and *Blocks World* probabilistic planning problems from the ICAPS 2004 IPPC [15]. In the *Box World* logistics problem, the domain objects consists of trucks, planes, boxes, and cities. The number of boxes and cities varies in each problem instance, but there were always 5 trucks and 5 planes. Trucks and planes are restricted to particular routes between cities in a problem-specific manner. The goal in *Box World* is to deliver all boxes to their destination cities, despite the fact that trucks and planes may stochastically fail to reach their specified destination. *Blocks World* is just a stochastic version of the standard domain where blocks are moved between the table and other stacks of blocks to form a goal configuration. In this version, a block may be dropped while picking it up or placing it on a stack.

We used the Vampire [20] theorem prover and the CPLEX 9.0 LP solver in our FOALP and FOAPI implementations and applied the basis function generation algorithm given in Fig. 1 to a FOMDP version of these domains. It is important to note that we generate our solution to the *Box World* and *Blocks World* domains offline. Since each of these domains has a universally quantified reward, our offline solution is for a generic instantiation of this reward. Then at evaluation time when we are given an actual problem instance (i.e., a set of domain objects and initial state configuration), we decompose the value function for each ground instantiation of the reward and execute a policy using the approach outlined in Sec. 6. We do not enhance or otherwise modify our offline solution once given actual domain information (this is an avenue for future research).

We set an iteration limit of 7 in our offline basis function generation algorithm and recorded the running times per iteration of FOALP and FOAPI; these are shown in Fig. 3. There appears to be exponential growth in the running time as the number of basis functions increases; this reflects the results of previous work [21]. However, we note that if we were not using the "orthogonal" basis function generation technique described in Sec. 5 and associated optimizations in Sec. 7, we would not get past iteration 2 of basis function generation due to the prohibitive amount of time required by FOALP and FOAPI ($> 10$ hours). Consequently, we can conclude that our basis function generation algorithm and optimizations have substantially increased the number of basis functions for which FOALP and FOAPI remain viable solution options. In terms of a comparison of the running times of FOALP and FOAPI, it is apparent that each performs better in different settings. In *Box World*, FOAPI takes fewer iterations of constraint generation than FOALP and thus is slightly faster. In *Blocks World*, the policies tend to grow more quickly in size because the Vampire theorem prover has difficulty refuting inconsistent partitions on account of the use of equality in this FOMDP domain. This impacts not only the solution time of FOAPI, but also its performance as we will see next.

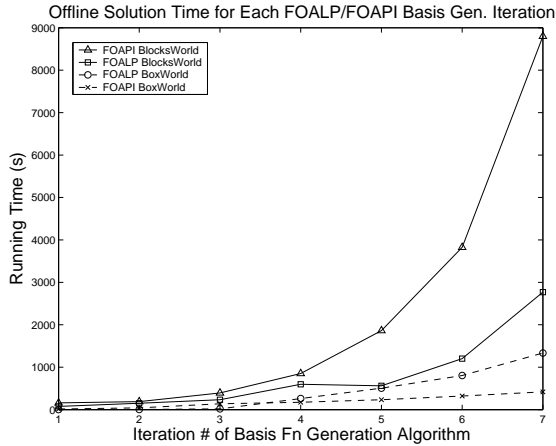

Figure 3: FOAPI and FOALP solution times for the Box World and Blocks World Domains vs. the iteration of basis function generation.

| Problem  | Prob. Planning System ||||| FO– ||
|----------|-----|-----|-----|-----|-----|-----|-----|
|          | G2  | P   | J1  | J2  | J3  | ALP | API |
| *bx c10 b5*  | 438 | 184 | 419 | 376 | 425 | 433 | 433 |
| *bx c10 b10* | 376 | 0   | 317 | 0   | 346 | 366 | 366 |
| *bx c10 b15* | 0   | –   | 129 | 0   | 279 | 0   | 0   |
| *bw b5*  | 495 | 494 | 494 | 495 | 494 | 494 | 490 |
| *bw b11* | 479 | 466 | 480 | 480 | 481 | 480 | 0   |
| *bw b15* | 468 | 397 | 469 | 468 | 0   | 470 | 0   |
| *bw b18* | 352 | –   | 462 | 0   | 0   | 464 | 0   |
| *bw b21* | 286 | –   | 456 | 455 | 459 | 456 | 0   |

Table 1: Cumulative reward of 5 planning systems and the FOALP and FOAPI (100 run avg.) on the *Box World* and *Blocks World* probabilistic planning problems from the ICAPS 2004 IPPC [15] (– indicates no data). *Box World* problems are indicated by a prefix of *bx* and followed by the number of cities *c* and boxes *b* used in the domain. *Blocks World* problems are indicated by a prefix of *bw* and followed by the number of blocks *b* used in the domain. All domains are available from *http://www.cs.rutgers.edu/~mlittman/topics/ipc04-pt*. See Section 8 for an explanation of the domains and planners.

We applied the policies generated by the FOALP and FOAPI versions of our basis function function generation algorithm to three *Box World* and five *Blocks World* problem instances from the ICAPS 2004 IPC. We compared our planning system to the three other top-performing planners on these domains: *G2* is a temporal logic planner with human-coded control knowledge [8]; *P* is an RTDP-based planner [2]; *J1* is a human-coded planner, *J2* is an inductive policy iteration planner, and *J3* is a deterministic replanner [24]. Results for all of these planners are given in Table 1.

We make four overall observations: (1) FOALP and FOAPI produce the same basis function weights and therefore the same policies for the *Box World* domain. (2) We only used 7 iterations of basis function generation and this effectively limits the lookahead horizon of our basis functions to 7 steps. It appears that a lookahead of 8 is required to properly plan in the final *Box World* problem instance and thus both FOALP and FOAPI failed on this instance.[5] (3) Due to aforementioned problems with the inability of FOAPI to detect inconsistency of policy partitions in the *Blocks World* domain, its performance is severely degraded on these problem instances in comparison to FOALP. FOALP does not use a policy representation and thus does not encounter these problems. (4) It is important to note that in comparing FOALP and FOAPI to the other planners, G2 and J1 used hand-coded control knowledge and J3 was a very efficient search-based deterministic planner that had a significant advantage because the probabilities in these domains were inconsequential. The only fully autonomous stochastic planners were P and J2, and FOALP performs comparably to both of these planners and outperforms them by a considerable margin on a number of problem instances.

---
[5]We could not increase the number of iterations to 8 to test this hypothesis due to memory constraints. We are currently working on additional optimizations to remedy this problem.

## 9 Concluding Remarks

We have introduced a novel algorithm for performing first-order approximate policy iteration, as well as new basis function generation techniques that allow FOALP and FOAPI to efficiently exploit their structure, leading to a substantial increase in the number of basis functions that these algorithms can handle. Additionally, we have addressed the intractability of solving problems with universal rewards by automatically decomposing the task into independent subgoals that can be solved and then recombined to determine a policy that facilitates "coordination" among the subgoals. Taken together, these techniques have enabled us to evaluate FOALP and FOAPI solutions to logistics problems from the ICAPS 2004 Probabilistic Planning Competition. Empirically we have shown that FOALP performs better than other autonomous stochastic planners on these problems and outperforms FOAPI when the pol-icy representation requires first-order logic constructs that pose difficulties for a state-of-the-art theorem prover. Our approach is competitive on these domains even with planners that exploit (hand-coded) domain-specific knowledge.

One pressing issue for future work is to extend our reward decomposition techniques to a wider range of universally quantified formulae. In addition, we note that many domains including the *Box World* logistics domain covered in this paper have an underlying topological structure that is not exploited in current solution algorithms. The ability to directly exploit topological structure in the problem representation and basis function formulation could potentially help with the limited-horizon lookahead issues that we experienced on *Box World*. The ability to exploit additional reward and domain structure will help push *fully lifted and automated first-order* solution techniques further into probabilistic planning domains that previously could not be handled.